\documentclass[runningheads]{llncs}
\usepackage[T1]{fontenc}
\usepackage{bbding}
\usepackage{graphicx}
\usepackage{amsmath,amssymb}
\usepackage{booktabs}
\usepackage[hidelinks]{hyperref}

\begin{document}

\title{Trust the Edge, Doubt the Arrow: Auditing Validation Signals for Text-Mined Clinical Causal Knowledge Graphs}
\titlerunning{Trust the Edge, Doubt the Arrow}

\author{Su Lan\raisebox{0.45ex}{*} \and
Shicheng Zhou\raisebox{0.45ex}{*} \and
Yiqi Wang \and
Yanming Zhu \and
Xuefei Yin \and
Alan Wee-Chung Liew\textsuperscript{\Envelope}}
\authorrunning{S. Lan et al.}
\institute{School of Information and Communication Technology, Griffith University,
Gold Coast, QLD, Australia\\
\raisebox{0.45ex}{*} These authors contributed equally to this work.\\
\textsuperscript{\Envelope} Corresponding author:
\email{a.liew@griffith.edu.au}.}

\maketitle

\begin{abstract}
Clinical causal retrieval-augmented generation systems increasingly use knowledge graphs (KGs) mined from notes, yet extracted assertions remain candidates that require audit for patient-level EHR support and direction. We construct a directed candidate graph from 328{,}839 MIMIC-IV discharge summaries using LLM extraction and UMLS normalization, and evaluate three signals against SemMedDB: within-admission co-occurrence \emph{lift}, cross-admission diagnosis-coding \emph{temporal precedence}, and \emph{literature corroboration}. Lift filters unsupported co-mentions and provides useful EHR support, but not causal proof. Coding order is a weak direction signal: among the 686 candidates where it commits, it reaches 54.2\% agreement with SemMedDB, below the extractor's 69.4\% on the same subset. Literature corroboration supplies independent direction evidence but covers only 30.4\% of candidate relations. A MIMIC-III replication yields the same pattern: extractor direction remains stable (70.5\%), whereas coding-order direction is near chance (50.9\%). Thus, text-mined clinical causal candidate KGs are best treated as \emph{EHR-supported but direction-uncertain}. In a leakage-controlled causal QA benchmark, lift-only retrieval achieves 35.7\% accuracy; conservative causal demotion achieves 37.2\% while reducing evidence-unsupported answers from 17.3\% to 6.1\% at fixed retrieval recall. We release code, aggregate summaries, and scripts to regenerate restricted outputs.

\end{abstract}

\section{Introduction}
Mining causal relations from clinical narratives (``hepatic encephalopathy \emph{due to} alcoholic cirrhosis'') can produce causal candidate KGs for clinical reasoning. Recent \emph{causal RAG} retrieves over text-mined causal graphs so that evidence carries direction~\cite{wang2025causalrag}. These systems must decide which extracted candidates are reliable enough to retrieve. That decision has two parts: whether an LLM-extracted causal assertion has patient-level EHR support, and whether its direction has independent support. An LLM reading notes can emit clinically plausible causal assertions, unsupported co-mentions (``chest pain $\rightarrow$ abdominal pain''), and reversed candidates. A natural audit strategy is to use structured EHR signals: keep candidate relations whose endpoints co-occur more than expected and orient them by the order in which diagnoses appear over time. Yet these assumptions are rarely tested against an external reference. Unsupported candidates add retrieval noise, while misdirected candidates can lead a system to answer a causal question with a downstream consequence rather than an upstream cause. We therefore ask which signals can support an extracted relation and which can provide usable direction evidence.

\begin{figure}[!htbp]\centering
\includegraphics[width=0.94\textwidth]{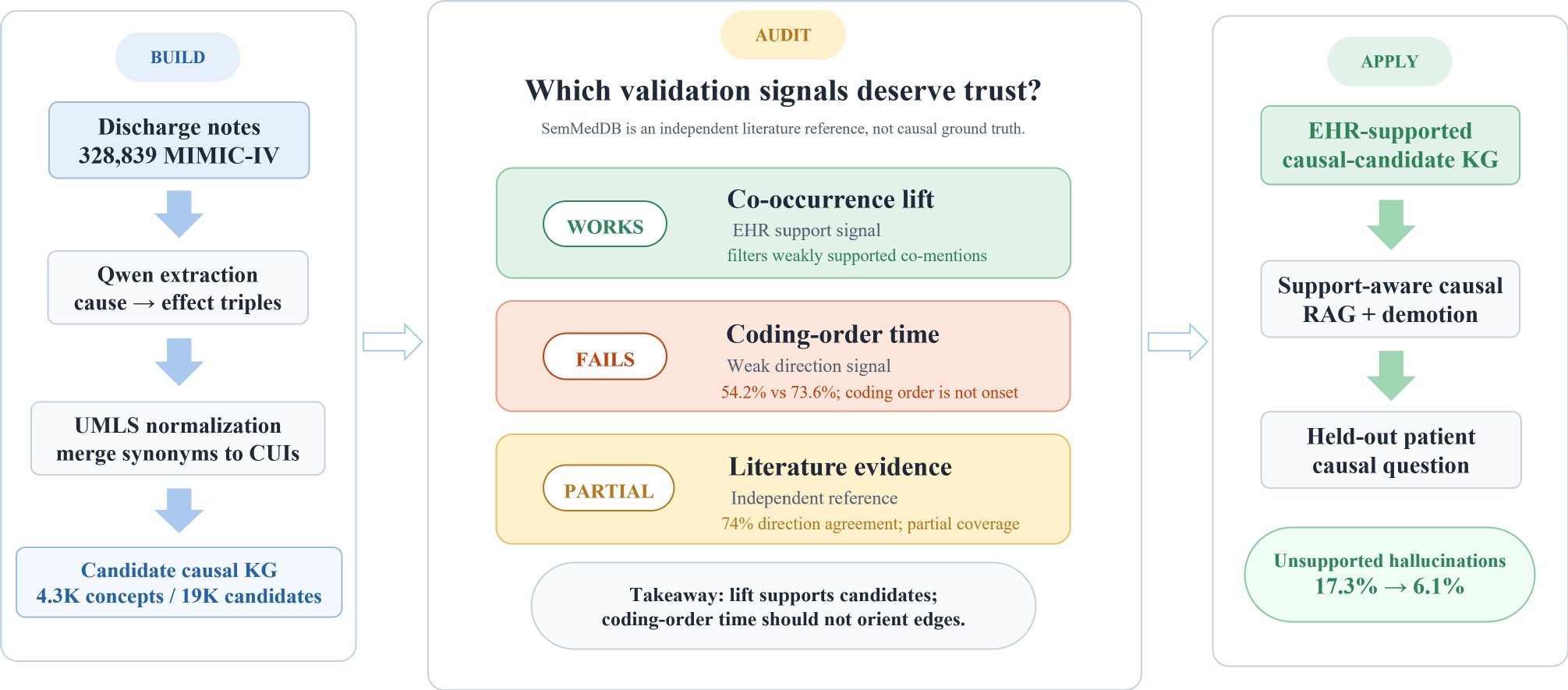}
\caption{Overview of the proposed audit-and-retrieval framework. \textbf{Build:} causal candidates are mined from clinical notes and normalized to UMLS concepts. \textbf{Audit:} co-occurrence lift provides EHR support, coding-order time is unreliable for direction, and SemMedDB provides partial literature corroboration. \textbf{Apply:} support-aware causal RAG retrieves from the EHR-supported candidate graph and demotes candidates that fail the causal screen, reducing evidence-unsupported answers at fixed recall.}
\label{fig:overview}
\end{figure}

We build a directed causal candidate KG from 328K MIMIC-IV discharge summaries and audit three validation signals against SemMedDB, a literature-derived causal-predicate database independent of our pipeline (Fig.~\ref{fig:overview}):
\begin{itemize}
\item \textbf{Co-occurrence lift} provides EHR support: it filters unsupported co-mention candidates and improves downstream causal-QA correctness.
\item \textbf{Cross-admission temporal precedence} is unreliable as an \emph{independent EHR direction signal}: against literature direction, it is correct on only 54.2\% of the candidates where it commits, worse than the extractor on the same subset (69.4\%). ICD coding order reflects \emph{discovery}, not \emph{onset}, a hazard known to structured causal discovery~\cite{shen2021causaldiscovery} that we quantify here for text-mined candidates.
\item \textbf{Literature corroboration} provides \emph{direction evidence but limited coverage}: it covers 30.4\% of all candidate relations and agrees with the extractor's direction on roughly three quarters of the covered subset.
\end{itemize}

Together, these results reveal a practical asymmetry: patient-level EHR support can be screened using a simple co-occurrence signal, whereas causal truth is not established and \emph{direction} remains uncertain. We therefore treat text-mined clinical causal candidate KGs as \emph{EHR-supported but direction-uncertain}: co-occurrence lift is used to filter unsupported candidates, the extractor's direction is retained as an imperfect text-derived estimate, and coding-order time is not used to orient candidates. Rather than proposing a new generator architecture, this work provides an audit and reusable benchmark for assessing when causal candidates mined from clinical text are reliable enough to support retrieval.

Our contributions are summarized as follows.
\begin{enumerate}
\item \textbf{A patient-grounded causal QA benchmark} with leakage-controlled, dual-signal reference answers. The benchmark pairs clinician-stated causal assertions with EHR support, making it a targeted test bed for support-filtered causal retrieval rather than a claim of causal ground truth.
\item \textbf{A systematic audit} of three validation signals against independent literature evidence, yielding the \emph{EHR-supported, direction-uncertain} asymmetry. Co-occurrence lift filters unsupported candidates, while cross-admission coding order is near chance and literature corroboration is coverage-limited.
\item \textbf{A reproducible pipeline for clinical causal candidate KG construction} from 328K discharge notes, combining LLM extraction, UMLS normalization, per-candidate audit metadata, and cross-release replication on MIMIC-III.
\item \textbf{A controlled causal-RAG study} in which lift-only retrieval serves as the audit-consistent baseline after the temporal-direction audit. A conservative causal-demotion policy preserves recall, substantially reduces evidence-unsupported answers, and gives a small non-significant accuracy trend.
\end{enumerate}


\section{Related Work}
\noindent\textbf{Clinical KGs from text and EHR.} Clinical KG methods commonly use ontologies, co-occurrence, or patient-specific associative graphs. GRAM~\cite{choi2017gram} and KAME~\cite{ma2018kame} use ICD ontologies; Rotmensch et al.~\cite{rotmensch2017learning} learn disease--symptom co-occurrence graphs; LLMs extract KGs from notes~\cite{nadkarni2023llmkg}; and GraphCare~\cite{jiang2024graphcare} builds personalized KGs for prediction. These graphs are mainly ontological or associative, not audited causal candidates.

\noindent\textbf{Causal KGs in medicine.} Biomedical causal KGs often use literature-scale predications such as SemMedDB~\cite{kilicoglu2012semmeddb}, combine literature relations with EHR evidence~\cite{nordon2019building}, or evaluate literature-derived graphs in EHR case studies~\cite{lyu2023dnkg}. Other work studies adverse drug effects~\cite{ade2026causalkg}, structured EHR causal discovery~\cite{shen2021causaldiscovery}, and causal GNNs~\cite{causalgnn2025survey}. A key challenge is validation: diagnosis timestamps often reflect documentation rather than onset, so naive temporal order can misorient relations~\cite{shen2021causaldiscovery,doutreligne2024causal}. We instead audit support and direction signals for causal candidates mined from patient narratives.

\noindent\textbf{KG-augmented prediction.} GraphCare~\cite{jiang2024graphcare}, KARE~\cite{kare2025}, RAM-EHR~\cite{xu2024ramehr}, MedText~\cite{lu2021medtext}, MM-STGNN~\cite{tang2023multimodal}, and KAT-GNN~\cite{katgnn2025} use graph structure to improve EHR outcome prediction. They optimize prediction rather than auditing whether extracted relations have EHR support or reliable direction.

\noindent\textbf{RAG, GraphRAG, causal RAG.} GraphRAG~\cite{edge2024graphrag}, Medical Graph RAG~\cite{medgraphrag2024}, and ontology-grounded clinical QA~\cite{llmkg2024biomed} retrieve KG evidence to reduce hallucination. CausalRAG~\cite{wang2025causalrag} retrieves over a text-extracted causal graph, so retrieved evidence carries direction. Causal-Audit~\cite{lan2026causalaudit} makes intervention reasoning auditable through target-aware causal-chain construction; our work addresses the complementary upstream problem of auditing clinical candidate-edge support and direction before retrieval. A raw causal-RAG variant is therefore used as a reference point in our evaluation. 

\noindent\textbf{Clinical QA benchmarks.} EHRNoteQA~\cite{kweon2024ehrnoteqa}, EHRXQA~\cite{ehrxqa2023}, EHRSQL~\cite{lee2022ehrsql}, and emrQA~\cite{pampari2018emrqa} mainly evaluate factual questions over notes, tables, or structured records. General-domain causal QA exists~\cite{yang2022causalreasoning}, but it is neither clinical nor patient-grounded. Our benchmark instead pairs patient-grounded causal questions with reference answers supported by causal statements in notes and structured EHR evidence.

\noindent\textbf{Positioning.} Table~\ref{tab:position} positions our work against representative systems along five axes. Existing clinical GraphRAG systems mainly use associative or ontology-based edges; CausalRAG studies directed retrieval outside patient EHR narratives; and literature-derived causal KGs provide causal predicates but are not mined from patient notes or audited for candidate direction. In contrast, our study jointly mines directed causal candidates from patient narratives, filters them with structured-EHR support, audits direction signals, and evaluates them with a patient-grounded causal QA benchmark. The study does not propose a new knowledge-graph-completion algorithm. It addresses the complementary quality-control problem that precedes completion and retrieval: determining which observed directed candidates have usable evidence support and which directions remain uncertain.

\begin{table}[!htbp]
\centering
\caption{Positioning against representative prior work. \checkmark{} indicates yes, $\sim$ partial, and $\times$ no. Ours is the only setting that combines directed causal candidate mining from patient narratives, structured-EHR support filtering, direction-signal auditing, and a patient-grounded causal QA benchmark.}
\vspace{-0.2cm}
\label{tab:position}
\setlength{\tabcolsep}{2pt}\footnotesize
\resizebox{\textwidth}{!}
{%
\begin{tabular}{lccccc}\toprule
 & Directed & Patient & EHR & Direction & QA \\
 & Candidates & Text & Support & Audited & Benchmark \\\midrule
GraphCare~\cite{jiang2024graphcare}, Med.\ GraphRAG~\cite{medgraphrag2024} & $\times$ & $\sim$ & $\times$ & $\times$ & $\times$ \\
CausalRAG~\cite{wang2025causalrag} & $\checkmark$ & $\times$ & $\times$ & $\times$ & $\times$ \\
Lit.\ causal KGs~\cite{kilicoglu2012semmeddb,nordon2019building,lyu2023dnkg,ade2026causalkg} & $\checkmark$ & $\times$ & $\sim$ & $\times$ & $\times$ \\
\textbf{Ours} & $\checkmark$ & $\checkmark$ & $\checkmark$ & $\checkmark$ & $\checkmark$ \\\bottomrule
\end{tabular}}
\vspace{-0.3cm}
\end{table}

\section{Building the Candidate Causal KG}
\noindent\textbf{Extraction.} 
A discharge summary mixes free-text clinical narrative with templated patient-instruction boilerplate (``call your doctor if\ldots''), which repeats across notes and can dominate raw edge frequency. We therefore restrict extraction to the History of Present Illness and Hospital Course sections, including ``Brief Hospital Course'', and prompt Qwen2.5-7B-Instruct~\cite{qwen2025} (AWQ, vLLM~\cite{kwon2023vllm}, temperature~0) to emit a JSON array of \texttt{\{"cause","effect"\}} pairs for each note. Over the 328{,}839 MIMIC-IV-Note discharge summaries~\cite{johnson2023mimicnote}, 79.5\% yield $\geq$1 triple, giving 1.31M triples (3.98 per note) and 1.17M unique surface edges. The unique-edge distribution is heavy-tailed: 96.4\% occur exactly once. Because the singleton tail is dominated by hyper-specific or noisy spans, the candidate KG retains only edges attested in $\geq$2 notes (42{,}798). Singleton extractions are excluded from the KG but retained as per-note provenance for benchmark grounding.

\noindent\textbf{Normalization.} 
Surface phrases contain synonyms, abbreviations, and modifiers, so we link each cause and effect span to a UMLS Concept Unique Identifier (CUI)~\cite{bodenreider2004umls} with a scispaCy~\cite{neumann2019scispacy} candidate generator. We accept the top candidate when cosine similarity is $\geq0.80$ and its semantic types are not in a non-clinical blocklist, thereby collapsing co-referring surface forms into concept nodes. Compared with a rule-based cue-phrase extractor, LLM extraction more than doubles the UMLS link rate (49.3\% vs.\ 21.9\%) and increases the giant connected component from 86.9\% to 97.1\% at comparable graph size (Table~\ref{tab:kg}), suggesting that LLM-extracted spans are more concept-like. Table~\ref{tab:kg} reports statistics on the full pre-split corpus. After the patient-level split used for leakage control (\S\ref{sec:qa}), the train-only KG used for retrieval and release contains 3{,}899 concept nodes and 16{,}273 directed edges.

\begin{table}[!htbp]\centering
\caption{Candidate-KG construction on the full pre-split corpus: rule-based vs.\ LLM extraction.}
\label{tab:kg}\footnotesize
\setlength{\tabcolsep}{4pt}
\begin{tabular}{lrr}\toprule
Metric & Rule-based & LLM(ours) \\\midrule
Raw surface edges & 142{,}890 & 1{,}173{,}644 \\
UMLS link rate & 21.9\% & \textbf{49.3\%} \\
CUI nodes & 8{,}572 & \textbf{4{,}318} \\
CUI edges & 19{,}171 & 19{,}035 \\
Giant component & 92.3\% & \textbf{97.1\%} \\
Lift-supported candidates & 736 & \textbf{2{,}047} \\\bottomrule
\end{tabular}\end{table}

\section{The Support and Direction Audit}\label{sec:audit}
We score three validation signals against two questions: whether an extracted causal candidate has patient-level EHR support, and whether its \emph{direction} has independent support. As an independent reference, we use SemMedDB v43~\cite{kilicoglu2012semmeddb}, a repository of 115.5M semantic predications extracted from PubMed. We keep the causal predicates \textsc{causes}, \textsc{predisposes}, and \textsc{induces}, read the subject$\rightarrow$object orientation as the literature's causal direction, and restrict the analysis to predications whose subject and object CUIs both occur in our KG. When both orientations are attested, we use the majority direction by citation support. This yields literature causal evidence for 4{,}943 KG candidate relations. Because SemMedDB is imperfect (relaxed $F_1\approx0.5$~\cite{kilicoglu2012semmeddb}), we treat it as an independent reference rather than ground truth, and compare each signal against both SemMedDB and the extractor's text-derived direction.

\smallskip\noindent\textbf{Experimental logic.} The audit and the downstream QA study answer different questions. The audit evaluates which validation signal can support candidate existence and which can recover direction. Each signal is reported by coverage and, when it commits to a direction, agreement with SemMedDB. The QA study then tests whether the useful part of the audit, support filtering by lift, improves causal RAG under a controlled benchmark. All QA systems share the same generator, prompt, and retrieval budget; differences therefore reflect the retrieved evidence rather than the surrounding model setup. 

\subsection{Co-occurrence lift provides EHR support}
We map ICD diagnoses to CUIs and, for candidate $A\rightarrow B$, compute the within-admission $\mathrm{lift}=P(B\mid A)/P(B)$. A candidate is treated as \emph{supported} when $\mathrm{lift}>1$, the positive-association boundary, with sufficient co-occurrence; 2{,}047 LLM candidates meet this criterion. Lift is \emph{symmetric}, so it measures association strength rather than direction or causal truth. It demotes weakly supported co-mentions: ``chest pain $\rightarrow$ abdominal pain'' (two common symptoms with high base rates) is not supported ($\mathrm{lift}\approx1$), whereas ``alcoholic cirrhosis $\rightarrow$ hepatic encephalopathy'' is strongly supported ($\mathrm{lift}=65$). Section~\ref{sec:qa} shows that retrieving over the lift-supported subgraph improves downstream causal-QA correctness. Co-occurrence lift is therefore useful for EHR support filtering, not causal proof or orientation.

\subsection{Cross-admission temporal precedence does not recover direction}
We hypothesized that across admissions a cause is coded \emph{before} its effect. We test this on the 4,943 candidate relations with SemMedDB direction evidence, comparing the extractor's text-derived direction with the direction implied by cross-admission coding order. As temporal precedence often abstains, we report both committed-subset accuracy and a coverage-adjusted paired comparison.

\begin{figure}[!htbp]\centering
\includegraphics[width=0.72\textwidth]{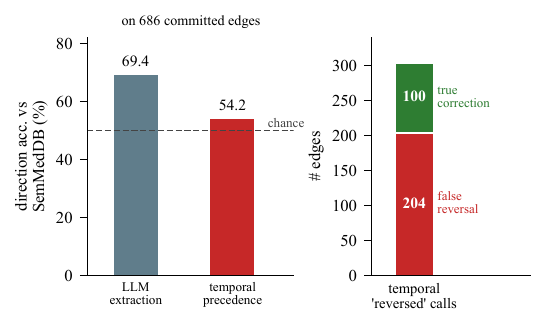}
\caption{Direction-signal audit. \emph{Left:} on the 686 candidate relations where cross-admission temporal precedence commits to a direction, it agrees with SemMedDB less often (54.2\%) than the extractor's own direction (69.4\%) and barely above chance. \emph{Right:} of temporal's 304 ``reversed'' verdicts, 204 are false reversals (coding artifacts) and only 100 are true corrections.}
\label{fig:audit}
\end{figure}

Table~\ref{tab:audit} and Fig.~\ref{fig:audit} show that temporal precedence is a weak direction signal. It commits on only 686 candidates and reaches 54.2\% agreement with SemMedDB, below the extractor's 69.4\% on the same subset. Over all 4,943 candidates, treating abstentions as failures to recover direction, the extractor is correct far more often than temporal precedence (LLM-only-right: 3,365; temporal-only-right: 100; McNemar $p \ll 0.001$). Fig.~\ref{fig:audit} further shows that most temporal reversals are harmful: among 304 reversed verdicts, only 100 are true corrections and 204 are false reversals.

\begin{table}[!htbp]\centering\caption{Direction-signal accuracy vs.\ SemMedDB ($n=4{,}943$ candidate relations). Temporal precedence abstains on 86\% of candidates; on the 686 where it commits it is below the extractor's own direction and near chance.}\label{tab:audit}\footnotesize
\begin{tabular}{lc}\toprule
Direction signal & Acc.\ vs.\ SemMedDB \\\midrule
Raw LLM extraction direction & \textbf{73.6\%} \\
Temporal precedence (committed, $n{=}686$) & 54.2\% \\
LLM direction on those 686 & 69.4\% \\\bottomrule
\end{tabular}\end{table}


\noindent\textbf{Interpretation.} 
ICD coding order reflects \emph{discovery/coding}, not \emph{onset}: chronic conditions are co-coded, and findings are coded at the encounter in which they are discovered. This timestamp hazard is known to structured causal discovery, where onset-correction (distinguishing incident from prevalent diagnoses) is the proposed remedy~\cite{shen2021causaldiscovery}. Here we quantify the same hazard for orienting text-mined causal candidates, where the temporal signal is barely above chance and below the extractor's own direction. SemMedDB itself has direction noise, but it and the extractor are both text-derived and agree on 73.6\% of all SemMedDB-covered directions, while the structured temporal signal correlates with neither. Applying onset-correction to the temporal signal is left to future work.

\noindent\textbf{Robustness.} 
The McNemar comparison is paired over all 4{,}943 SemMedDB-covered candidates. Each candidate has an extractor verdict, and the temporal signal is counted as correct only if it commits to the literature direction. Abstentions are therefore failures for an independent directioner. The committed-only accuracy in Table~\ref{tab:audit} is a diagnostic subset, not the McNemar denominator. The gap is unlikely to be a SemMedDB artifact. The extractor and SemMedDB are produced by independent text-mining pipelines yet agree on 73.6\% of all SemMedDB-covered directions, whereas the structured temporal signal agrees with neither. This is the pattern expected if diagnosis-coding order carries little causal-direction information. The conclusion is also stable under stricter literature-reference thresholds. When requiring at least 1, 3, or 10 supporting PubMed records per SemMedDB relation, temporal committed-set agreement remains 54.2\%, 55.3\%, and 53.8\%, respectively, while extractor agreement on the corresponding literature subsets is 73.6\%, 76.1\%, and 79.3\%.

\subsection{Literature corroboration is useful but partial}
Among the 2{,}047 lift-supported candidates, 886 (43\% of the lift-supported subset) are also attested by SemMedDB. The attested subset agrees with the extractor's direction in 74\% of cases and covers 61\% of benchmark reference relations. Its value depends on how it is used. As a strict filter requiring both EHR-lift support and literature attestation, it would discard the remaining 1{,}161 lift-supported EHR candidates (57\%) that fall outside SemMedDB's PubMed-derived coverage. This would hurt recall. As a confidence tier, it separates high-confidence candidates (lift \emph{and} literature) from EHR-only candidates and supplies an independent direction for the subset it covers. We therefore use it as coverage-limited corroboration rather than as a strict filter.

\smallskip\noindent\textbf{Takeaway.} Co-occurrence lift provides patient-level EHR support for extracted candidates; \emph{direction} remains uncertain because coding-order time is near chance and literature covers a minority of candidates. Text-mined clinical causal candidates are therefore best treated as EHR-supported but direction-uncertain. We filter by lift, keep the extractor's direction as an uncertain estimate, and avoid orienting candidates by coding-order time.

\subsection{Putting it together: a support-filtering recipe}
\begin{figure}[!htbp]\centering
\includegraphics[width=0.66\textwidth]{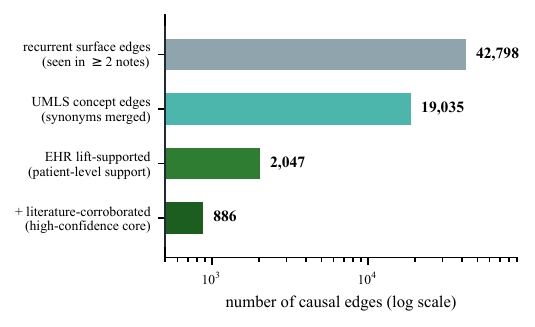}
\caption{The support-filtering recipe as a funnel. Recurrent LLM candidates are normalized to UMLS concepts, filtered to the EHR lift-supported subgraph, and partitioned by literature corroboration into a higher-confidence core; direction is taken from the extractor rather than coding-order time.}
\label{fig:funnel}
\end{figure}

The audit yields a reproducible recipe for converting raw LLM-extracted causal candidates into an EHR-supported candidate graph, without manual curation (Fig.~\ref{fig:funnel}). Starting from 42{,}798 recurrent surface candidate relations, we first normalize endpoints to UMLS concepts (\S\ref{sec:audit} reduces 16{,}151 surface nodes to 4{,}318 concepts). We then keep candidates with within-admission lift~$>1$, yielding 2{,}047 supported candidate relations, and orient each candidate by the extractor's direction rather than by coding-order time. Finally, we flag the 886 literature-corroborated candidates as a higher-confidence core and treat the remainder as EHR-only. The resulting graph has patient-level support annotations grounded in text and structured co-occurrence, without establishing causal truth. Its \emph{direction} claims retain source-aware uncertainty. Section~\ref{sec:qa} shows that this graph supports downstream causal QA retrieval, and the recipe transfers to note corpora paired with linked structured records.

\subsection{Cross-release replication on MIMIC-III}
\begin{figure}[!htbp]\centering
\includegraphics[width=0.72\textwidth]{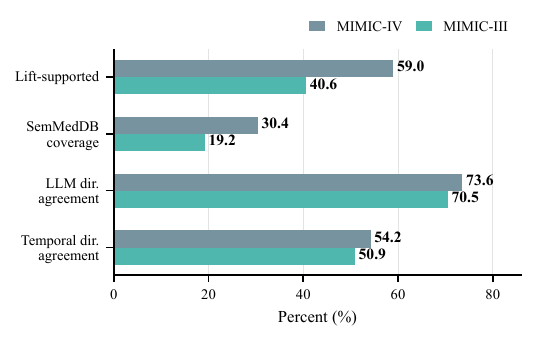}
\caption{Cross-release replication on MIMIC-III. Compared with the main MIMIC-IV audit, MIMIC-III has lower lift support and lower SemMedDB coverage, but the direction pattern is stable: the extractor's direction remains substantially above chance, whereas coding-order direction remains near chance.}
\label{fig:mimic3}
\end{figure}

To test whether the audit depends on the MIMIC-IV release, we reran the same extraction, UMLS normalization, structured-diagnosis lift support filtering, temporal analysis, and SemMedDB comparison on MIMIC-III. The replication uses 53{,}981 usable discharge-summary sections and yields 237{,}957 clean triples, 221{,}454 surface edges, and 35{,}232 CUI-level edges. Although this graph is larger and noisier than the MIMIC-IV train graph, the qualitative conclusions replicate (Fig.~\ref{fig:mimic3}): 942 of 2{,}318 diagnosis-testable candidates are lift-supported (40.6\%); SemMedDB covers 6{,}753 candidates (19.2\%); extractor direction remains aligned with SemMedDB (70.5\%, close to the 73.6\% overall MIMIC-IV extractor agreement); and coding-order direction remains near chance (50.9\%, close to the 54.2\% MIMIC-IV committed-subset result). Thus MIMIC-III strengthens the central audit claim under a different EHR release: the EHR support pattern transfers, while naive temporal orientation remains unreliable.

\section{Causal QA Benchmark and Downstream Utility}\label{sec:qa}
\subsection{Benchmark}
\smallskip\noindent\textbf{Reference construction.}
A candidate relation becomes a benchmark answer only under a \emph{dual signal}. It must be stated as a cause and effect relation by the treating clinician in a \emph{held-out} note, and it must be EHR-lift-supported ($\mathrm{lift}\geq2$). This support threshold is stricter than the graph-construction threshold ($\mathrm{lift}>1$): the graph threshold keeps useful retrieval candidates, whereas the benchmark threshold reduces weak associations in the reference set. This pairing removes symptom co-mention noise that either signal alone would admit. 

\smallskip\noindent\textbf{Leakage control.}
We split the 145{,}872 patients of MIMIC-IV~\cite{johnson2023mimiciv} 85/15 by \texttt{subject\_id} (124{,}058 train / 21{,}814 test). The candidate KG is built from \emph{train} notes only, so no test patient's note contributes a candidate relation. Population-level lift may use all patients' codes, a different modality that cannot reveal an individual's text answer. Questions are drawn from \emph{test} patients and answered against the train-only candidate KG. Because lift also defines our support-filtered retrieval condition, this benchmark is a downstream utility check under the same operational support definition, not an independent proof that lift establishes causal truth.

\smallskip\noindent\textbf{Statistics and evaluation.} The pool has 2{,}319 items; downstream evaluation uses 466 WHY and 466 WHATCAUSES questions over 170 supported candidates. Direction-discrimination templates are reserved for future work. Each question includes the patient's coded problem list, and in $\sim$half the reference cause is absent from that list. A balanced audit sample contains 40 items per template. Manual review of visible fields and clinical direction marked 102 pass, 15 flag, and 3 uncertain; source notes were not re-read, so this audits benchmark construction rather than the automatic judge. Correctness requires a reference CUI match or Qwen confirmation of the same condition; faithfulness requires the answer to appear in retrieved evidence. We call an answer evidence-unsupported when it does not appear in that evidence; this is a retrieval-faithfulness error and does not imply that the answer is clinically false.

\subsection{Does EHR support filtering help causal RAG?}
\noindent\textbf{Why these baselines.} No published system exactly matches patient-grounded causal QA over an EHR-mined causal candidate graph under this support-filtering setting. Rather than compare against systems that also change the generator, chunking, or prompt, we re-implement retrieval paradigms under one harness. All systems share the same generator (Qwen2.5-7B), prompt, and retrieval budget, and differ \emph{only} in the retrieved evidence. The baselines are: (1)~\textbf{closed-book}, none; (2)~\textbf{naive TF-IDF text-RAG}, top-3 discharge-note excerpts; (3)~\textbf{SemMedDB-only}, a directed literature causal graph for external biomedical KG grounding~\cite{kilicoglu2012semmeddb}; (4)~\textbf{assoc-graph}, undirected co-occurrence neighbours in a GraphRAG-style associative retrieval setting~\cite{edge2024graphrag,medgraphrag2024,jiang2024graphcare}; (5)~\textbf{raw-causal}, the raw LLM directed candidate graph with no support filtering, as a CausalRAG-style reference~\cite{wang2025causalrag}; (6)~\textbf{lift + temporal}, the lift-supported graph after applying the audited coding-order temporal direction signal; (7)~\textbf{lift-only}, the EHR-supported candidate graph in the extractor's original direction; and (8)~\textbf{lift-only + causal demotion} (ours), which keeps the same top-12 lift candidates but moves candidates that fail the causal screen to the end of the evidence list. We additionally evaluate causal-cue TF-IDF and MiniLM sentence retrieval as stronger text-retrieval checks under the same harness; their overall scores are reported in Section~\ref{sec:analysis}. Thus, differences reflect the retrieval source or evidence ordering rather than changes to the generator, prompt, or retrieval budget. 
Raw-causal directly ablates support filtering. Lift + temporal shows the cost of using the rejected coding-order direction signal: recall falls to 77.8\%. Lift-only directly ablates causal demotion.

\smallskip\noindent\textbf{Retrieval setup.}
For the graph systems, retrieval is direction-aware: a WHY question pulls the \emph{in}-edges of the asked effect (its candidate causes) and a WHATCAUSES question its \emph{out}-edges, ranked by lift, frequency, or PMID support. The top~12 are serialized into the prompt, and the lift-only systems additionally annotate each candidate with its lift. The generator returns a single condition, which we then match. Generation and judging run on a 7B open model on a private cluster, so MIMIC text does not leave the compliant environment.

\smallskip\noindent\textbf{Causal-screen demotion.} The demotion layer is a post-retrieval ordering rule, not a new retriever. Lift-only retrieval first selects the same top-12 candidates by co-occurrence lift. The causal screen is queried only for those directed pairs: for WHY, candidate cause$\rightarrow$queried effect; for WHATCAUSES, queried cause$\rightarrow$candidate effect. The screen is estimated on the training split as a new-user, incident-outcome design over structured diagnoses: the cause must occur at baseline, the effect must first appear later among outcome-naive patients, and each patient needs follow-up ($\geq$2 admissions). The adjusted risk-ratio signal uses inverse-propensity weighting for age, sex, and baseline comorbidity burden, then calibrates a $z$ score with automatically sampled negative-control diagnosis outcomes that the cause does not point to in the candidate KG; these controls are not manually chosen and do not use held-out QA answers. The implementation samples up to 20 controls per tested pair, requires at least five valid controls for calibration, and estimates a pair only when both exposure arms have at least 50 eligible patients and at least 10 incident outcome events. A candidate \emph{passes} if $z\geq1.64$, \emph{fails} if it is estimated but below that one-sided 5\% threshold, and is \emph{untested} otherwise. The cutoff, screen, and QA prompts are fixed before answer generation and are not tuned on held-out QA answers. Demotion moves failed candidates to the end of the same top-12 list while leaving pass and untested candidates in lift order; it changes ordering but cannot add or remove candidates.

\smallskip\noindent\textbf{Causal-screen coverage.}
Table~\ref{tab:screen} summarizes how often the causal screen affects the lift-only evidence lists. 
\begin{table}[!b]\centering
\caption{Causal-screen coverage and ranking effect in the lift-only top-12 retrieval lists. The screen is applied after lift-only retrieval and only reorders the selected candidates. Lower mean rank indicates earlier placement in the evidence list.}
\label{tab:screen}
\scriptsize
\setlength{\tabcolsep}{3pt}
\begin{tabular}{lrrrrr}\toprule
Screen status & Slots & \% & Unique pairs & Qs $\geq$1 & Mean rank before$\to$after \\\midrule
Pass ($z\geq1.64$) & 1{,}718 & 18.6 & 127 & 630 (67.6) & 4.63$\to$3.78 \\
Fail ($z<1.64$) & 4{,}510 & 48.7 & 494 & 807 (86.6) & 7.09$\to$8.39 \\
Untested & 3{,}025 & 32.7 & 377 & 780 (83.7) & 5.54$\to$4.08 \\
Total & 9{,}253 & 100 & -- & 932 & -- \\\bottomrule
\end{tabular}
\end{table}
Across the 932 questions, lift-only retrieval produces 9,253 top-12 candidate slots. The screen assigns 18.6\% of slots to pass, 48.7\% to fail, and leaves 32.7\% untested because of sample-size or coverage requirements. Demotion changes the evidence order in 678 questions (72.7\%) without changing the retrieved candidate set, so recall is preserved by construction. Failed candidates move down the list on average, while pass and untested candidates move upward relative to them.

\begin{table}[!htbp]\centering
\caption{Causal QA on held-out patients (\%). $n=932$ paired. Faith. denotes answer faithfulness to retrieved evidence; Unsup. denotes evidence-unsupported answers.}
\label{tab:qa}\footnotesize
\setlength{\tabcolsep}{3.5pt}
\resizebox{\textwidth}{!}{%
\begin{tabular}{lcccccc}\toprule
System & WHY & WHATCAUSES & Overall & Faith. & Unsup. & Recall \\\midrule
1~Closed-book & 22.5 & 9.0 & 15.8 & N/A & N/A & N/A \\
2~Naive TF-IDF text-RAG & 18.2 & 7.5 & 12.9 & N/A & N/A & N/A \\
3~SemMedDB-only & 24.9 & 13.5 & 19.2 & 83.8 & 16.2 & 23.9 \\
4~Assoc-graph & 35.4 & 16.3 & 25.9 & 73.9 & 26.1 & 72.7 \\
5~Raw-causal & 35.8 & 18.7 & 27.3 & 83.2 & 16.8 & 72.3 \\
6~Lift + temporal & 47.2 & 24.0 & 35.6 & 86.3 & 13.7 & 77.8 \\
7~Lift-only & 46.4 & 25.1 & 35.7 & 82.7 & 17.3 & \textbf{86.3} \\
\textbf{8~Lift-only + demotion (ours)} & \textbf{48.5} & \textbf{26.0} & \textbf{37.2} & \textbf{93.9} & \textbf{6.1} & \textbf{86.3} \\\bottomrule
\end{tabular}}\end{table}

\begin{figure}[!htbp]\centering
\includegraphics[width=0.86\textwidth]{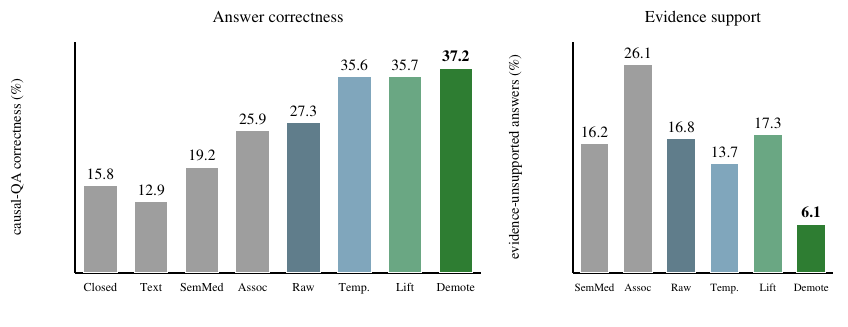}
\caption{Causal-QA performance on held-out patients. Left: answer correctness. Right: evidence-unsupported answers for systems where evidence support can be measured. The final system uses lift-only retrieval and demotes candidates that fail the causal screen. The main downstream effect is fewer evidence-unsupported answers at fixed recall; the exact-accuracy change over lift-only is a small non-significant trend.}
\label{fig:qa}
\end{figure}

\smallskip\noindent\textbf{Results.}
Table~\ref{tab:qa} reports the held-out causal-QA results, and Fig.~\ref{fig:qa} visualizes the main trends in answer correctness and evidence support. In the broad comparison, graph-based retrieval outperforms closed-book generation and the implemented naive TF-IDF text-RAG baseline. The support-aware comparison is the cleaner ablation after the audit: lift-only retrieval avoids the refuted temporal flip, and causal demotion preserves the same retrieval recall while raising the measured correctness scores. The overall gain over lift-only is modest (+1.5 points; McNemar $p=0.27$), so we do not claim a statistically significant accuracy improvement. The robust effect is faithfulness: evidence-unsupported answers fall from 17.3\% to 6.1\% at the same 86.3\% retrieval recall.

\subsection{Analysis of the gain}\label{sec:analysis}
This experiment tests retrieval utility rather than causal truth. The question fixes direction, lift-only retrieval avoids unsupported temporal reorientation, and demotion only reorders the same top-12 candidates. The result is therefore improved evidence discipline: faithfulness rises and the evidence-unsupported answer rate falls at fixed recall, while exact-concept accuracy changes non-significantly.

\smallskip\noindent\textbf{Recall is preserved.}
Lift-only retrieval contains the reference answer for 86.3\% of items, compared with 72.3\% for raw-causal, 72.7\% for the associative graph, and 23.9\% for SemMedDB-only. Demotion preserves 86.3\% recall because it only reorders the same candidate set.

\smallskip\noindent\textbf{Question-type effects.} WHATCAUSES remains harder than WHY (26.0\% vs.\ 48.5\% for our final system), reflecting its one-to-many answer space and the benchmark's single-reference design. Demotion improves both question types over lift-only retrieval.

\smallskip\noindent\textbf{Error analysis.} Residual errors mainly involve clinically plausible sibling concepts penalized by exact-CUI matching, retrieval gaps ($\sim$22\% of items), and salient but incorrect concepts selected from noisy evidence. A fully UMLS-aware semantic evaluation is not reported. Stronger sanity checks within the same harness also remain below closed-book: causal-cue TF-IDF reaches 13.9\% overall and MiniLM sentence retrieval 13.1\%.

\section{Discussion and Limitations}
\noindent\textbf{Practical recommendations.} The audit yields four practical rules. First, use within-admission co-occurrence lift as an EHR support filter before retrieval. Second, do not orient candidates by diagnosis-coding order. Third, if a richer causal screen is available, use it to demote failed candidates rather than to replace the retrieval list. Fourth, treat literature corroboration (e.g.\ SemMedDB) as a confidence tier with limited coverage.

\noindent\textbf{On the temporal hazard.} Coding timestamps often reflect documentation rather than onset, a known problem in structured causal discovery~\cite{shen2021causaldiscovery}. We quantify this hazard for text-mined candidates: the naive coding-order signal agrees with neither the extractor nor the literature reference. This does not reject all temporal methods. It only shows that uncorrected cross-admission coding order is a weak independent direction signal.

\noindent\textbf{Threats to validity.} The QA reference is lift-supported by construction, so the downstream comparison is coupled to the deployed support definition rather than an independent causal validation of lift. The raw-causal baseline contains a superset of these candidates and therefore remains the cleaner ablation of unsupported-neighbour removal. Population lift also uses all patients' structured codes; a train-only estimate would be stricter. We report literature-reference sensitivity but not lift-threshold or causal-screen-cutoff sweeps, and stronger BM25, domain-specific dense, and hybrid text retrieval remain unevaluated. The lift and causal-screen cutoffs are pre-specified operational choices rather than clinically validated optima. The 120-item manual audit checks visible benchmark fields without re-reading source notes, and the Qwen judge is not calibrated against clinician-annotated system outputs. Results use one 7B model and English ICU notes. Lift is limited to diagnosis-coded endpoints; generalisation to medications, procedures, laboratory findings, and other clinical entity types is not established. SemMedDB has partial coverage, and neither association signal establishes causal truth. Accordingly, the strongest supported downstream conclusion is a lower evidence-unsupported answer rate at fixed recall, not a significant accuracy gain or clinical causal validation.

\noindent\textbf{Released artifacts.} The \href{https://github.com/VincentZ007/clinical-causal-kg-artifact}{public repository} provides code, aggregate results, and the MIMIC-III summary. Restricted MIMIC, UMLS, and SemMedDB materials are not redistributed; credentialed users can regenerate the KG, benchmark, and audit labels.

\section{Conclusion}
Text-mined clinical causal candidate KGs require separate audits of support and direction. Lift provides EHR support but not causal proof; coding order is unreliable for direction; and literature corroboration is coverage-limited. This pattern replicates on MIMIC-III, supporting an EHR-supported but direction-uncertain interpretation. In causal-RAG evaluation, conservative demotion reduces evidence-unsupported answers from 17.3\% to 6.1\% at fixed recall, while accuracy changes non-significantly. Support signals should therefore be audited before they are used to support, orient, or rank candidates.

\subsubsection*{Disclosure of Interests.}
The authors have no competing interests to declare that are relevant to the content of this article.

\begingroup
\let\oldthebibliography\thebibliography
\let\endoldthebibliography\endthebibliography
\renewenvironment{thebibliography}[1]{%
  \oldthebibliography{#1}%
  \footnotesize
  \setlength{\itemsep}{0pt}%
  \setlength{\parsep}{0pt}%
}{\endoldthebibliography}
\bibliographystyle{splncs04}
\bibliography{references}
\endgroup
\end{document}